\documentclass[lettersize,journal]{IEEEtran}
\usepackage{graphicx}
\usepackage{cite}
\usepackage{picinpar}
\usepackage{url}
\usepackage{flushend}
\usepackage[latin1]{inputenc}
\usepackage{colortbl}
\usepackage{soul}
\usepackage{multirow}
\usepackage{pifont}
\usepackage{color}
\usepackage{alltt}
\usepackage[hidelinks]{hyperref}
\usepackage{enumerate}
\usepackage{siunitx}
\usepackage{breakurl}
\usepackage{amsmath,amssymb,amsfonts}

\begin{document}
\title{An Integrated Hardware-Software Design for Low-Data Spatial Defect Detection in Robotic Visual Inspection with Hybrid Optoelectronic Neural Networks}

\author{
	\vskip 1em
	
	Chaoqing Tang, 
	Jiaxuan Li, 
         Huanze Zhuang,
         Guiyun Tian, 
         Chao Wang, 
         Yihao Ouyang,
	   and Wenzhong Liu

	 \thanks{


        
	 	Chaoqing Tang, Jiaxuan Li, Huanze Zhuang, Yihao Ouyang and Wenzhong Liu are with School of Artificial Intelligence and Automation, Huazhong University of Science and Technology, Wuhan, 430074, China (e-mail: billtang@hust.edu.cn)
        
             Chaoqing Tang and Wenzhong Liu are also with China Belt and Road Joint Lab on Measurement and Control, Wuhan, 430074, China.

             Guiyun Tian is with School of Electric and Electrical Engineering, Chongqing University of Technology, Chongqing, 400054, China

             Chao Wang is with Department of Precision Instrument, Tsinghua University,  Beijing, 100084, China
            
	 }
}

\maketitle
	
\begin{abstract}
To address the challenges of data bandwidth overload and inefficient shape-level annotation in robotic visual inspection systems, this paper proposes an optoelectronic architecture that integrates hardware and software. This non-imaging, low-data paradigm drastically reduces annotation requirements. First, adopting a sensor-in-the-loop strategy, a Digital Micromirror Device (DMD) is reconfigured as a physical optical convolutional layer, enabling feature extraction directly in the photonic domain to unify sensing hardware with feature extraction software. To control data volume at the source, a block-based compressed sensing strategy is designed, encoding spatial information into low-dimensional temporal signals, thereby significantly reducing data redundancy. Subsequently, to circumvent inefficient manual defect shape annotation, natural language descriptions of defect shapes are utilized to guide the neural network in aligning with highly generalizable features from Contrastive Language-Image Pre-training (CLIP). This successfully steers the attention maps of the optoelectronic neural network to approximate defect shapes. Finally, we propose the Localization Accuracy for Attention (LAA) metric to quantify the shape-level defect localization capability of attention heatmaps. Validation experiments on transparent material defect detection demonstrate the system's effectiveness. Parametric analysis reveals the impact of measurement matrices, compression ratios, and block sizes on accuracy. Results show that compared to traditional imaging methods, the proposed architecture maintains parity in accuracy while reducing data volume by 90\% for Vision Transformers and computational workload by 60\% for Convolutional Neural Networks. This low-data paradigm provides an effective solution for industrial automation scenarios characterized by massive data streams, high acquisition costs, or constrained edge resources.
\end{abstract}


\markboth{ }%
{}

\definecolor{limegreen}{rgb}{0.2, 0.8, 0.2}
\definecolor{forestgreen}{rgb}{0.13, 0.55, 0.13}
\definecolor{greenhtml}{rgb}{0.0, 0.5, 0.0}

\section{Introduction}
\label{sec:introduction}
\IEEEPARstart{S}{patil} image-based defect detection has been extensively deployed in various automated industrial robot systems \cite{RN4, RNa1, RNa2}, encompassing fields such as machining \cite{RN17} and aerospace \cite{RN5}. In these applications, inspection workflows predominantly rely on high-definition (HD) images. While HD imaging enables precise capture of minute defect features, it inevitably induces exponential growth in data volume. For instance, high-resolution images used in large liquid crystal panel inspections can generate datasets exceeding several gigabytes \cite{RN19}. Such massive data scales pose significant challenges to storage efficiency and computational processing capabilities, particularly for industrial edge computing environments. Moreover, certain industrial defect detection applications, such as laser/infrared scanning imaging, incuring excessive time consumption during image acquisition. At the algorithmic level, deep learning-based object detection frameworks (e.g., You Only Look Once (YOLO) \cite{RN29, RNa3}) dominate spatial image defect detection. However, these methods heavily depend on extensive labeling efforts for target localization and classification \cite{RN32}, demanding substantial time and labor costs. Furthermore, ensuring the accuracy and consistency of labeling results remains a persistent challenge.

To address the challenge of massive data volumes, both industry and academia have proposed various processing methods. Image resizing techniques reduce data size by lowering image resolution; while straightforward to implement, this approach sacrifices fine-grained details. In the context of tiny defect detection, low-resolution images may fail to clearly exhibit defect characteristics, leading to frequent missed or false detections that undermine high-precision requirements. When applying YOLO for object detection, inputting high-resolution images (e.g., 4K or 8K) typically triggers model-driven downsampling prior to processing, which significantly increases the missed detection rate. For instance, if retaining subtle defect features necessitates a 4K resolution, resizing such an image to 320$\times$320 pixels would render YOLO incapable of detecting targets occupying less than 2\% of the total image area \cite{RN27}.

The emergence of Compressed Sensing (CS) \cite{RN7} offers a novel sensor-in-loop and hardware-level approach to address the challenge of high data volume. By transcending the limitations of the traditional Nyquist sampling theorem, this technique enables signal acquisition at sampling rates significantly lower than the Nyquist rate and reconstructs images through specific algorithms. Theoretically, it reduces data volume by orders of magnitude while preserving resolution. This allows spatial image signals to be converted into time-domain sequences \cite{RNa4} with substantially reduced data size, demonstrating potential value across diverse fields such as single-pixel imaging \cite{RN930, RN1288}, stochastic optical reconstruction microscopy \cite{RN364}, and action inference \cite{RN15}. However, practical applications of CS still encounter bottlenecks. Conventional CS methods require complex reconstruction processes that are computationally expensive and time-consuming \cite{RN10}. Some studies have attempted direct target classification using CS data \cite{RN22, RN20}; while these bypass the reconstruction step, they remain heavily dependent on labeled data and suffer from low accuracy and an inability to localize defects. Furthermore, separate networks must be trained for different sampling rates or matrices, limiting model generalization and increasing training costs. Although some reconstruction-free classification methods that embed sampling information have improved accuracy to a certain extent, they still fail to achieve defect localization \cite{RN1}. Another category of CS-based detection methods involves first reconstructing the spatial image and then employing models like YOLO for classification and localization \cite{RN12}. These approaches not only require positional and categorical labels for training but also suffer from error accumulation during reconstruction, which degrades detection accuracy. Simultaneously, the data volume input to neural networks remains largely unoptimized.

In the software side, training-free large models based on vision-language cross-modal alignment have garnered significant attention to mitigate the heavy reliance on extensive annotated data \cite{RN13, RN25, RN24}. Methods such as CLIP \cite{RN14} map image regions and textual semantics into a unified embedding space, enabling direct matching between visual features and novel categories without additional training, thereby maintaining robust accuracy for unknown targets. Representative works include: YOLO-World \cite{RN25}, which dynamically converts text embeddings via the RepVL-PAN module and achieves 35.4 mAP on LVIS zero-shot detection, balancing open-world generalization and real-time efficiency; UniDetector \cite{RN24}, which integrates multi-source heterogeneous label spaces and decouples detection stages, capable of detecting over 7000 categories; and the Agentic Object Detection framework \cite{RN21} open-sourced by Andrew Ng's team, which accomplishes precise defect identification and localization without annotations through iterative agent optimization. However, since these models are trained on generic object names, their accuracy drops significantly in specialized industrial scenarios involving professional defect terminologies, such as impact damage on transparent materials. Moreover, they still rely on high-resolution inputs and fail to address the fundamental data volume challenge posed by high-definition imaging.

In summary, due to the decoupling between data acquisition and processing workflows, as well as between hardware and software, current image-based spatial defect detection systems fail to simultaneously achieve label-free operation and low-data-volume requirements. Regarding data-task coupling, C. Tang et al. \cite{RN918} proposed an integrated data acquisition framework. This framework aligns specific task processing demands with data collection to balance volume against detection objectives. In terms of hardware-software collaborative detection, Lin et al. introduced the all-optical diffractive deep neural network in 2018 \cite{RN28}. This method utilizes programmable multilayer diffractive surfaces to construct ``optical neurons." It enables light-speed inference and energy-efficient computation in the terahertz regime using light-based hardware for processing. To adapt to mature silicon-based electronic semiconductor technologies, opto-electronic hybrid neural networks have emerged. These networks seamlessly integrate with mainstream architectures (e.g., CNNs, Transformers) while leveraging light-speed latency and ultra-low energy consumption. For instance, Zhou et al. \cite{RN30} proposed a reconfigurable diffractive processing unit. This unit merges the high-throughput computing capability of optical diffraction with electronic programmability. Consequently, inspired by opto-electronic hybrid neural networks, this paper aims for hardware-software co-design. 

Therefore, to address the challenges of high data volume for current image-based methods and the need for extensive shape-level annotations in industrial defect detection robotic systems, inspired by opto-electronic hybrid networks, compressive sensing, and training-free large models, the key innovations are as follows:
\begin{enumerate}
    \item This paper proposes an integrated hardware-software optoelectronic architecture with unified data acquisition and processing, aiming to significantly reduce both data volume and end-to-end processing time in spatial defect detection. Compared to current image-based architectures, the proposed method achieves a 90\% reduction in data redundancy while maintaining accuracy and efficiency.
    \item The proposed method integrates pre-trained vision-language models with a novel defect localization strategy, reducing the requirement for both categorical and morphological-level annotations in traditional morphological defect detection to merely the preparation of defect category image datasets. Specifically, if data are acquired via factory-provided defect samples, no annotation is required, whereas if utilizing existing image datasets, only categorical labels are necessary. To accommodate this new localization approach, a novel evaluation metric termed Localization LAA is proposed to quantify this capability.
    \item We analyzed key parameters in this method through experiments on tiny spot and scratch defects detection in transparent materials, such as measurement matrix type, popular classification neural networks, compression rate and block size.
\end{enumerate}


\section{The Proposed Methods}
\subsection{Overall Diagram}
The core concept of this work is, first, to establish sensing-computing integration at the hardware physical layer. Specifically, a DMD is reconfigured as the first optical convolutional kernel, deeply merging the CS measurement process with feature extraction. This enables partial signal feature extraction and dimensionality reduction to occur directly within the optical signal domain. Consequently, data volume is reduced greatly, and the time-consuming image reconstruction process inherent in traditional CS is completely eliminated. Second, at the software algorithm level, natural language descriptions of defect shapes are utilized with the pre-trained vision-language (e.g., CLIP) model to generate robust shape features. These features guide the neural network to automatically achieve precise alignment between the feature space and defect shapes during training, allowing attention maps (e.g., via CAM) to accurately depict defect shapes. Therefore, compared to decoupled frameworks like YOLO that require extensive precise bounding boxes and categorical labels, the proposed method enables annotation-free operation in scenarios where factories provide defective samples, while simultaneously achieving shape-level defect localization.

\begin{figure}
    \centering
    \includegraphics[width=\linewidth]{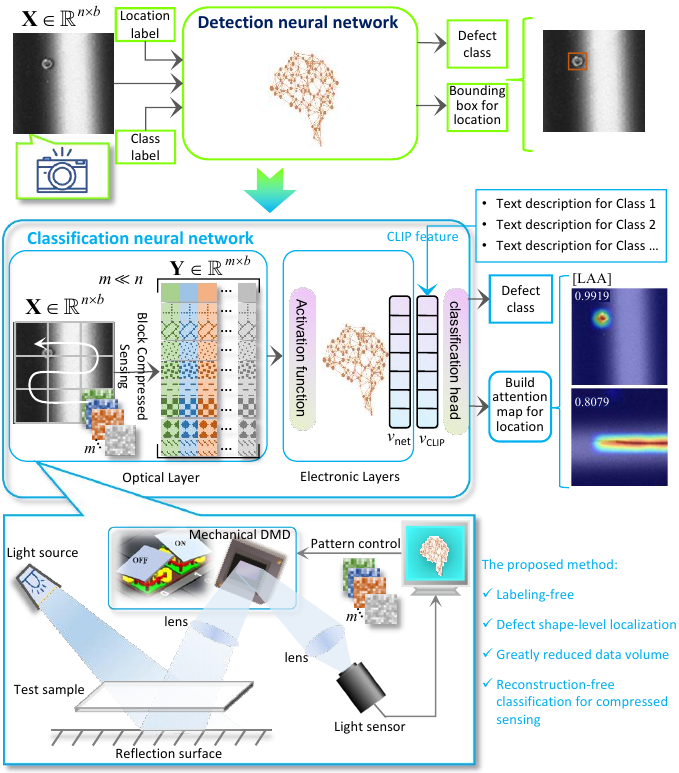}
    \caption{The overall diagram of the proposed optoelectronics architecture for robotic visual defect inspection. Traditional and the proposed method are highlighted in green and blue boxes, respectively. The implementation of optical layer is designed for double transmission-based detection materials like transparent materials, it works for reflection model too.}
    \label{fig1}
\end{figure}

Based on the core concepts, the architecture proposed in this paper consists of the following three main components, as illustrated in Fig.~\ref{fig1}:

\begin{enumerate}
\item Compressed Physical Encoding: Target field-of-view (FOV) compression is performed based on the structure of the first layer of artificial neural networks, such as CNNs, Recurrent Neural Networks, and Transformers. This paper focuses on convolution operations for spatial defect detection.
\item Annotation-free Shape Feature Learning: The pre-trained vision-language model is employed to map natural language descriptions into visual defect shape features, which supervises the backbone network.

\item Attention Map for Shape-level Localization: High-resolution attention maps are generated based on the trained neural network to achieve border-free, shape-level defect localization.
\end{enumerate}

Detailed implementations of each component will be introduced in the subsequent subsections.

\subsection{Compressed Physical Encoding with Block Compressed Sensing}
For a signal
$\mathbf{x}\in {{\mathbb{R}}^{n\times 1}}$, CS measurement process can be represented as:
\begin{equation}
    \mathbf{y=\Phi x}+\mathbf{\xi }
\end{equation}
where $\mathbf{\Phi }\in {{\mathbb{R}}^{m\times n}}$, $\mathbf{y}\in {{\mathbb{R}}^{m\times 1}}$ are the measurement matrix and measured data, respectively. $\mathbf{\xi }\in {{\mathbb{R}}^{m\times 1}}$ is the sampling noise. The original signal $x$ is compressed from $n$ to $m$ due to $m\ll n$. As is shown in Fig.\ref{fig1}, an image is evenly represented as $b$ blocks, and each block has $n$ pixels. If the image size is unable to evenly devided, perform zero-padding. $m$ measurement matrices are used to slidely get the CS measurement with dot product, so each block will get $m$ values, obtaining a compressed matrix with size $m\times b$ from the original image size of $n\times b$. This sliding measurement process can be regarded as a convolution layer in neural networks, where $\mathbf Y$ is the output feature map of convolution. 

One physical implementation for the above measurement process for transparent materials is given in the bottom box in Fig.\ref{fig1}. The light source is projected through the test sample, then focused onto a DMD. The DMD can independently control each micromirror at a flipping frequency of up to 200 kHz, with the number of mirrors reaching 4K resolution (4096$\times$2176). A pre-prepared measurement matrix is loaded by the computer to control the ON/OFF state of each pixel on the DMD. Finally, the light that carries defect information are focused on to  photoelectric sensor via a lens to obtain $\mathbf Y$. The sliding window process is implemented by sequentially activate a window on DMD. Since the photoelectric sensor measures the total light intensity of each projected frame, a spatial image is converted into a continuous time-domain measurement sequence. For reflective detection scenarios involving opaque materials, this method also works.

In terms of $\mathbf{\Phi }$, random Gaussian matrices and 0/1 Bernoulli matrices are commonly used. In this paper, considering physical implementation constraints, we employ the first-layer convolutional kernels from the trained neural network as $\mathbf{\Phi }$. 

\subsection{Annotation-free Shape Feature Learning with Pre-trained Vision-language Model}
The raw physical measurements $\mathbf Y$ are first sum up along the $m$-dimension, generating $b$ features per image. After activation, this feature map is functionally equivalent to the output of a single convolutional layer. To capture spatial dependencies, we utilize mainstream backbone networks, i.e., CNNs or Vision Transformers (ViTs), to encode these features into an output vector $v_{net}$.

\begin{figure}
    \centering
    \includegraphics[width=\linewidth]{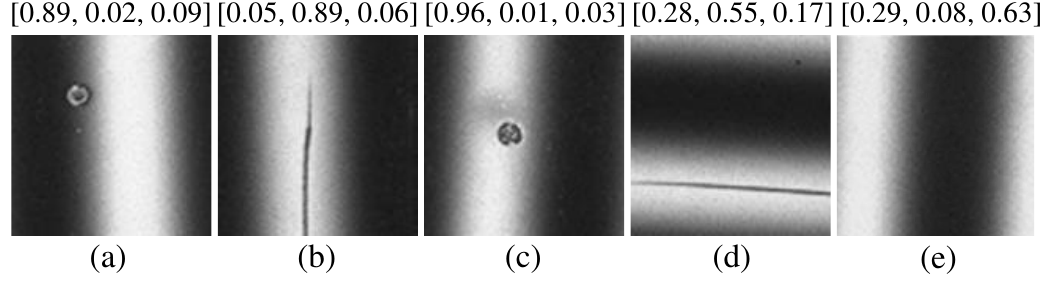}
    \caption{The classification probability for some examples with the designed class description texts in CLIP. The number in ``[ ]" indicates the probability for spot defect, scratch and health, sequentially.}
    \label{fig2}
\end{figure}

Instead of relying on laborious manual annotations for diverse defect shapes, this work exploits the rich cross-modal knowledge embedded in pre-trained models like CLIP. We use the feature embeddings of morphological descriptions as supervisory signals, thereby eliminating the need for explicit defect labels. In the case study application of this paper, i.e. tiny defect detection on transparent plates, three classification texts are designed to effectively distinguish the class of spot defect, scratch, and health, i.e. ``a photo with large circular speckle shape", ``a photo with slim long line shape scratch", ``a photo without obvious speckle or long line shape". Fig.~\ref{fig2} demonstrates that the classification probabilities align with human intuition, where the ground-truth class receives the highest score.

To implement shape feature learning, the output feature $v_{net}$ should be the same size with CLIP feature $v_{CLIP}$, normally 512 for most pre-trained models. So, the training loss function can be simply:
\begin{equation}
    L=mse\left( {{v}_{net}},{{v}_{CLIP}} \right)
\end{equation}
where $mse(\cdot)$ is the the mean square error.

\subsection{Attention Map for Shape-level Localization}
Since deep features inherently align with defect shape during training, this study employs attention maps for precise defect localization. A heatmap, originally a neural network visualization technique, essentially represents the weights indicating the specific regions of the input data that the network focuses on. In our detection framework, by upsampling the feature layer weights to match the FOV, we can highlight regions critical for classification, thereby achieving shape-level defect localization. This approach is particularly effective for sparsely distributed defects, such as those encountered in industrial inspection scenarios with good yield rates. Conversely, attention maps tend to disperse when defects are densely clustered within the FOV. Furthermore, as these maps are derived directly from pre-trained network weights and current inputs, the method eliminates the need for additional positional annotations, significantly reducing the labor cost associated with shape-level labeling required by traditional algorithms like YOLO.

Commonly utilized heatmap generation techniques include Grad-CAM, Score-CAM, and EigenCAM. Given that shallow neural network layers often introduce excessive nuisance details whereas deeper layers capture more generalized semantic features, we select the final feature extraction layer for analysis. Specifically, we utilize the last convolutional layer for CNNs and the final attention module layer for ViTs. Through comparative experiments, we observe that EigenCAM outperforms alternative methods in precisely localizing sparsely distributed defects (e.g., subtle cracks or small speckles in industrial inspection). It achieves superior focus on defect regions while effectively suppressing background noise.

\subsection{Metric of Localization Accuracy for Attention (LAA)}
\begin{figure}
    \centering
    \includegraphics[width=1\linewidth]{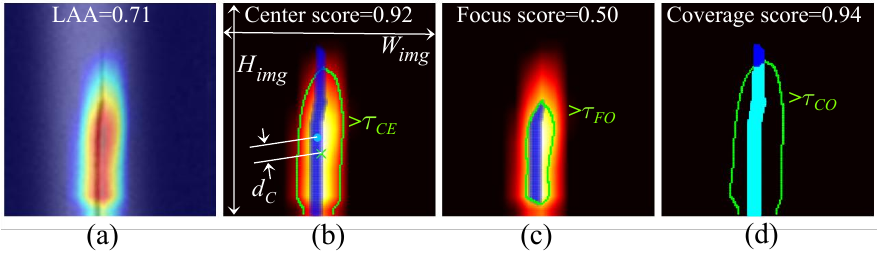}
    \caption{Explanation for how (a) LAA is calculated with the process of (b) center score, (c) focus score, and (d) coverage score using a demo image in Fig. \ref{fig2}(b). The blue section in (b-d) are the location label, and the heatmap within the green edges are the key attention part under their threshold values.}
    \label{fig3}
\end{figure}

Traditional bounding box-based localization methods use Intersection over Union (IoU) as a metric, but IoU cannot capture the morphological and attentional properties of heatmaps. Therefore, this paper proposes a metric called LAA. LAA evaluates localization precision via the center score ($S_{CE}$), attention focus via the focus score ($S_{FO}$), and ground truth coverage via the coverage score ($S_{CO}$). The overall calculation for LAA is given in Eq.(\ref{eq3}), and explanations are provided in Fig. \ref{fig3}. All scores range from 0 to 1.

\begin{equation}
    LAA={\left( {{S}_{CE}}+{{S}_{FO}}+{{S}_{CO}} \right)}/{3}
    \label{eq3}
\end{equation}

As is shown in Fig. \ref{fig3}, the center score is defined in Eq.(\ref{eq4}),
\begin{equation}
    {{S}_{CE}}=1-{{{d}_{C}}}/{\max \left( {{H}_{img}},{{W}_{img}} \right)}
    \label{eq4}
\end{equation}
where $d_C$ is the distance for weighted center. The weighted center for the ground truth label and attention map are denoted as blue solid circle and green cross in Fig.\ref{fig3}(b), respectively. $H_{img}$ and $W_{img}$ are the height and width of the original image, respectively. The weighted center coordinate ($C_{x,y}$) for heatmap is defined as Eq.(\ref{eq5}), 
\begin{equation}
    {{C}_{x,y}}=\sum\limits_{i}{{{w}_{i}}\cdot {{\hbar }_{i}}}=\sum\limits_{i}{{{{e}^{10{{I}_{A\left| {{\hbar }_{i}} \right.}}}}}/{\sum{{{e}^{10{{I}_{A\left| \hbar  \right.}}}}}}\;\cdot \hbar }
    \label{eq5}
\end{equation}
where $I_A$ is the attention image, ${{\hbar }_{i}}$ is the $i$-th coordinate in X or Y axis for $I_A$ where the value is greater than a threshold $\tau_{CE}$. Since the label image has only binary value, the center coordinates for the ground truth label can be calculated by simply apply average for all coordinates for the label.

The focus score is defined in Eq.(\ref{eq6}). Firstly, accumulating the attention values that greater than a threshold $\tau_{FO}$ while intersecting with the ground truth, $I_G$ is a binary label image which has `1' for the ground truth. Next, focus score is calculating the ratio of the accumulated value to the summation of all attention values that greater than the $\tau_{FO}$. This score represents the percentage of significant attention region in the ground truth label area.
\begin{equation}
    {{S}_{FO}}={\sum{{{I}_{G}}\cdot {{I}_{A\left| {{I}_{A}}>{{\tau}_{FO}} \right.}}}}/{\sum{{{I}_{A\left| {{I}_{A}}>{\tau_{FO}} \right.}}}}
    \label{eq6}
\end{equation}

The coverage score is defined as Eq.(\ref{eq7}), which calculates the number of intersection pixels between the ground truth label and the attention map where the values are greater than a threshold $\tau_{CO}$ first, and divided by the total number of `1' in $I_G$. So, coverage score evaluates the percentage of ground truth labels in the the attention region. Users can define their own threshold values for $\tau_{CE}$, $\tau_{FO}$, $\tau_{CO}$, we recommend setting them as 0.3, 0.8 and 0.2 respectively and empirically for general purpose. The Python code to calculate LAA will be available here\footnote{www.LAAtestwebsite.com} soon.
\begin{equation}
    {{S}_{CO}}={\sum{{{I}_{G}}\cdot B_A }}/{\sum{{{I}_{G}}}}
    \label{eq7}
\end{equation}

\section{EXPERIMENTAL SETTINGS}
\subsection{Tiny Defect Detection Systems for Transparent Materials}
\begin{figure}
    \centering
    \includegraphics[width=1\linewidth]{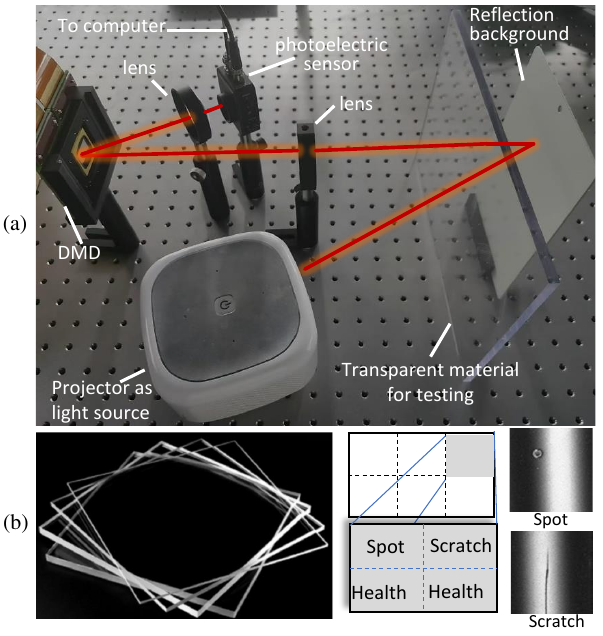}
    \caption{The (a) experimental setups and (b) specimens for tiny defect detection on transparent materials.}
    \label{fig4}
\end{figure}

This study was validated using a tiny defect detection system for transparent materials. As shown in Fig. \ref{fig4}(a), the system employs sinusoidal structured light illumination to enhance imaging contrast of tiny defects in transparent materials, thereby improving detection capability. The core detection optical path aligns with the transmission-reflection configuration illustrated in Fig. \ref{fig1}. For non-transparent materials, this optical path remains applicable and does not necessarily require structured light illumination (e.g., coaxial illumination can be used instead). Specifically, the structured light illumination in this experiment was generated by a Changhong Q2 Pro projector with a resolution of 1920$\times$1080. The structured light patterns were pre-generated on a computer and fed into the projector. The DMD (Vialux 9501) comprises a 1920$\times$1080 micromirror array and operates at a binary mode frequency of 17,857 Hz. The photodetector (PDA100A2) has a spectral response range of 320 to 1100 $nm$ and a maximum frequency response of 11 MHz.

As shown in Fig.~\ref{fig4}(b), experimental validation utilized transparent polyvinyl chloride (PVC) plates (300 $mm$$\times$200 $mm$) with eight thickness gradients (i.e., 1, 2, 3, 4, 5, 6, 8, and 10 $mm$). Each plate was divided into 2$\times$3 regions, where spot and scratch defects were prefabricated alongside defect-free controls. Defects were realistically simulated: spots via random falling objects and scratches via controlled tooling. This resulted in 2$\times$3$\times$8=48 independent regions per defect type, yielding approximately 200 samples per category.

\subsection{Experimental Settings}
In the proposed optoelectronic architecture, the DMD physically implements convolutional kernels. As shown in Fig.~\ref{fig1}, the binary nature of micromirrors necessitates specific mechanisms for four parameter types: Signed (-1 to 1), Unsigned (0 to 1), Bool (0 or 1), and Sign (1 or -1). For Unsigned types, the photodetector operates in integration mode, where grayscale control is achieved by duty-cycle modulation, which will limit the sampling speed. For example, the DMD used in this experiment operates at only 266 Hz in grayscale mode. For Bool types, the photodetector works in instantaneous mode, where the DMD directly outputs 1 or 0 via micromirror switching. For Signed types, parameters can be decomposed into positive+zero and negative+zero components, both treated as Unsigned cases. The negative+zero component is then inverted and summed with the positive+zero result. Similarly, Sign parameters are processed by separating 1 and -1 components.

This study experimentally analyzes key factors governing classification and localization accuracy: convolutional parameter types, CS sampling rates ($S_r$), and block sizes within the FOV. System parameters, including sinusoidal structured light patterns and component distances, are calibrated according to \cite{RN33}, as this work prioritizes computational methods over optical mechanism optimization. Training data consist of full-field images captured under structured light illumination across varying thicknesses and defect types. During this phase, the DMD is set to an all-ON state and the photodetector is replaced by an industrial camera. Thanks to the spatially distinct fabrication and imaging of defect samples, the dataset is pre-classified upon acquisition, eliminating manual labeling. The final dataset comprises 1,618 impact damage images, 1593 scratch images, and 1637 defect-free images. For applications with existed image datasets, simple categorical partitioning suffices, incurring workload comparable to class-only annotation. To validate performance, 20\% of the dataset was randomly selected and manually annotated with defect shapes via LabelMe. During deployment, the camera is replaced by the photodetector, and synchronization is initiated using short, periodic full ON-OFF-ON DMD sequences. Model training employs the Adam optimizer, with learning rates iteratively tuned to maximize accuracy.
 
\section{Results and Discussions}
\subsection{Classification Accuracy under Different Compression Rates}
\begin{figure}
    \centering
    \includegraphics[width=1\linewidth]{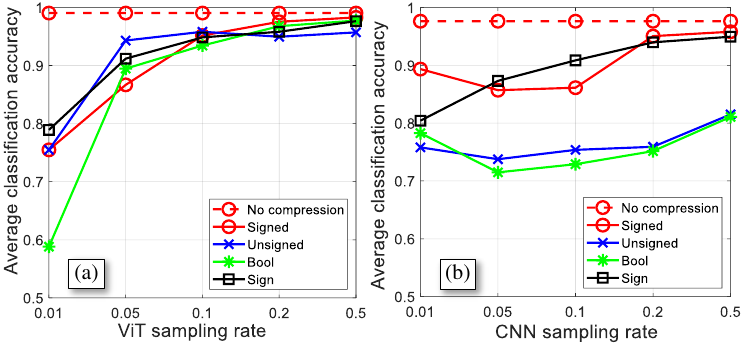}
    \caption{Average classification accuracy of (a) ViT and (b) CNN vs. sampling rate under different types of measurement matrix, where block size is set as 16. No compression refers to image-based method.}
    \label{fig5}
\end{figure}

This section evaluates the relationship between classification accuracy, sampling rate, and sampling matrix types. The comprehensive average classification accuracies for spot defects, scratches, and healthy samples under both ViT and CNN architectures are illustrated in Fig. \ref{fig5}, the image-based method is denoted as `No compression'. Given the simplicity of this proof-of-concept scenario, the ViT architecture employs only three attention layers, while the CNN consists of four convolutional feature extraction layers followed by three fully connected classification layers, augmented with batch normalization and dropout for enhanced performance. The first convolutional layer in both networks is implemented optically. Overall, classification accuracy improves with increasing sampling rates for both CNN and ViT, with ViT achieving superior performance due to its inherent alignment with the image-blocking strategy proposed in this work. At a sampling rate of $S_r=0.1$, ViT attains classification accuracies of approximately 0.95 across all matrix types, only 3 percentage points lower than the uncompressed case, which reduces 90\% of data). In the CNN backbone, Signed and Sign measurement matrix types exhibit significantly higher accuracy, whereas ViT shows negligible differences among matrix types. This suggests that ViT can leverage Bool-type matrices (physically faster to implement), drastically accelerating computation in the optical neural network layer.

\subsection{Classification Accuracy under Different Block Sizes}
\begin{figure}
    \centering
    \includegraphics[width=1\linewidth]{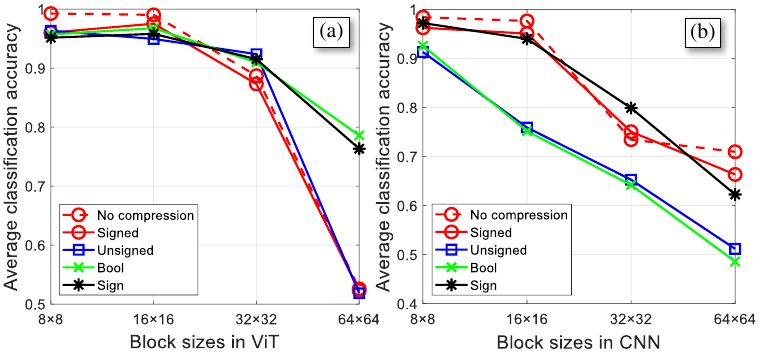}
    \caption{Average classification accuracy of (a) ViT and (b) CNN vs. block size under different types of measurement matrix, where $S_r$ is set as 0.2. No compression refers to image-based method.}
    \label{fig8}
\end{figure}

Block size is another critical factor influencing data compression rates. While ViT typically employs a block size of 16$\times$16, this section explores sizes ranging from 8$\times$8 to 64$\times$64. When the block size is 8$\times$8 with $S_r=0.01$, the parameter $m$ reaches its minimum value of 1. Fig.~\ref{fig8} illustrates classification accuracy versus block size under a fixed sampling rate of 0.2. The results first show that larger blocks degrade accuracy for both ViT and CNN. This degradation occurs because feature maps derived from the physical layer scale to ${\mathbb{R}^{1 \times b}}$ after summation along the m-dimension. Increased block size reduces $b$, creating an information transmission bottleneck. Secondly, ViT consistently outperforms CNN under identical conditions, maintaining an accuracy of approximately 0.9 at a block size of 32. Thirdly, in CNNs, signed kernels exhibit superior performance, whereas ViT shows no significant difference except at a block size of 64. At this scale, binary matrices (Bool/Sign) achieve markedly higher accuracy than multi-value types. The first potential reason is that increased spatial complexity within larger blocks (e.g., 64$\times$64) enhances the sparse filtering of key features by binary matrices, suppressing redundant noise. In contrast, multi-value matrices introduce high-frequency noise or interference via continuous values. The second reason is that ViT's self-attention mechanism prioritizes global semantic relationships over local details. This mechanism synergizes with coarse-grained binary features but conflicts with fine-grained multi-value details. Consequently, multi-value details disperse attention weights and reduce robustness.

\subsection{Localization Accuracy}
\begin{figure}
    \centering
    \includegraphics[width=1\linewidth]{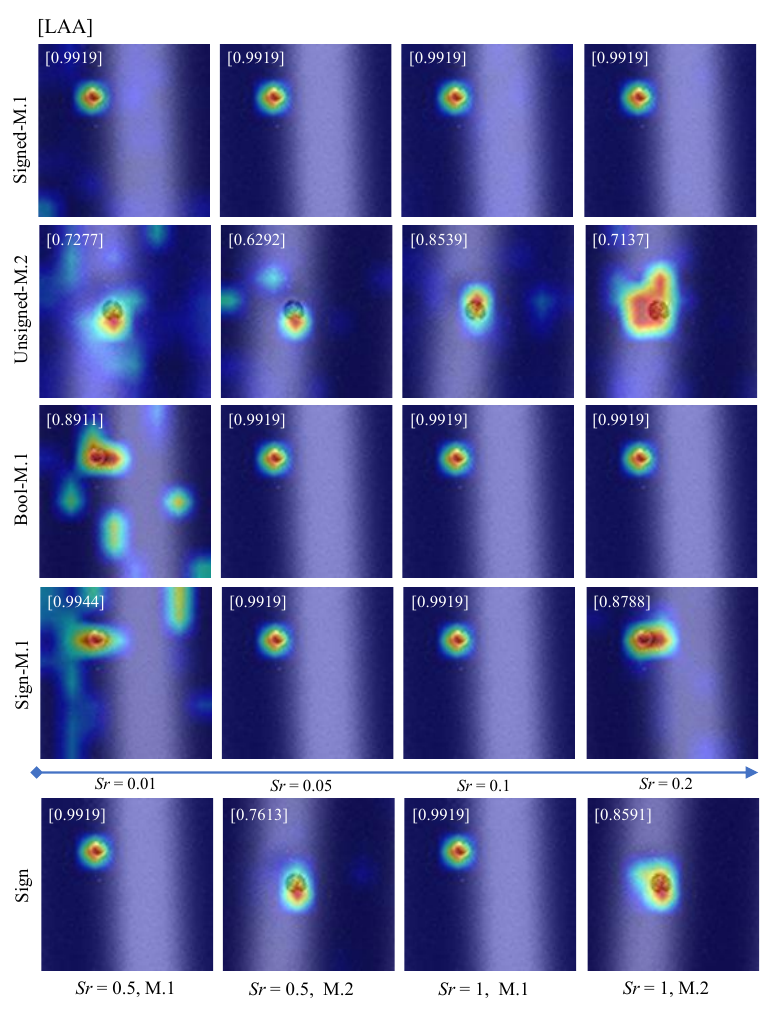}
    \caption{Examples for localization with attention map for ViT on two spot defects (M.1 and M.2), where block size is 16.}
    \label{fig11}
\end{figure}

\begin{figure}
    \centering
    \includegraphics[width=1\linewidth]{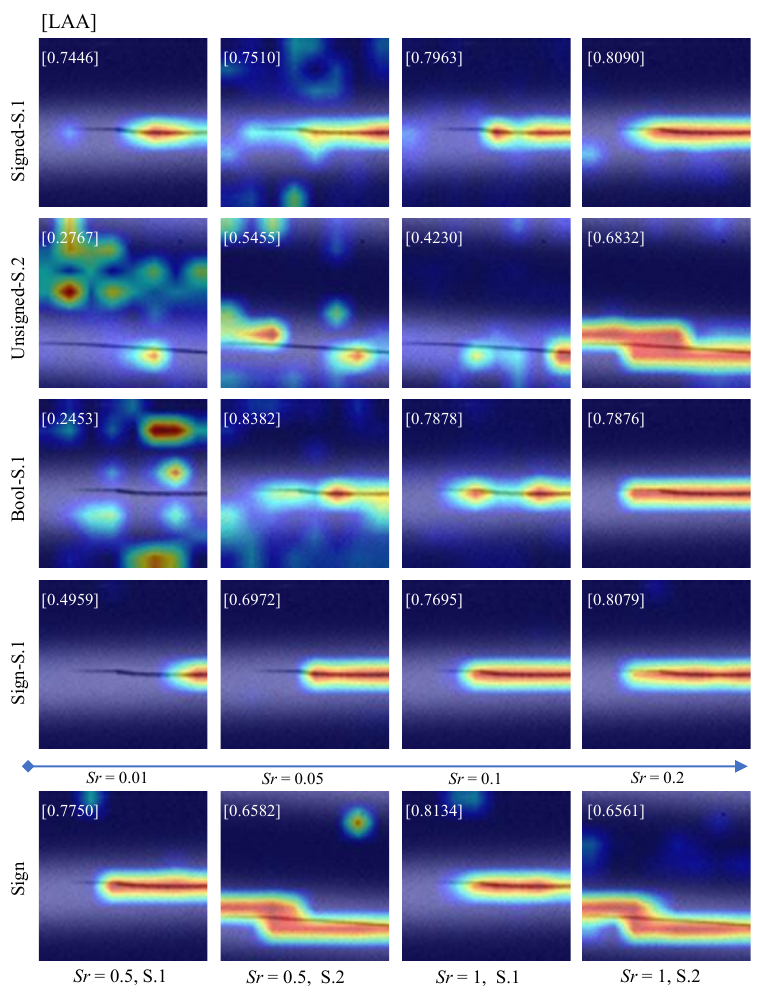}
    \caption{Examples for localization with attention map for ViT on two scratches (S.1 and S.2), where block size is 16.}
    \label{fig12}
\end{figure}

The localization performance of representative attention maps for both defect types is illustrated in Fig.~\ref{fig11} and Fig.~\ref{fig12}. Overall, increasing the sampling rate enhances both classification and localization accuracy. On the other hand, under identical sampling rates, spot defects achieve higher localization accuracy than scratches, which is consistent with the trend observed in classification accuracy. Given that the best LAA value for scratches in Fig.~\ref{fig12} does not exceed 0.9, the default LAA threshold may be overly stringent for scratches, and users may adjust it as needed.

These results confirm a strong correlation between the proposed LAA metric and visual localization quality. High LAA values (e.g., exceeding 0.9 in Fig.~\ref{fig11}) occur only when the center distance, attention focus, and ground truth coverage are simultaneously optimized. As shown in the latter two subfigures of the Unsigned-M.2 row in Fig.~\ref{fig11}, although the center distances are nearly identical, the LAA value drops to 0.7137 due to excessive dispersion of the key regions. Furthermore, a comparison of the first row in Fig.~\ref{fig12} (LAA = 0.7446 vs. LAA = 0.8090) demonstrates that, given similar center distances and dispersion levels, more comprehensive coverage of the ground truth labels by key regions leads to higher LAA scores.


\begin{figure}
    \centering
    \includegraphics[width=1\linewidth]{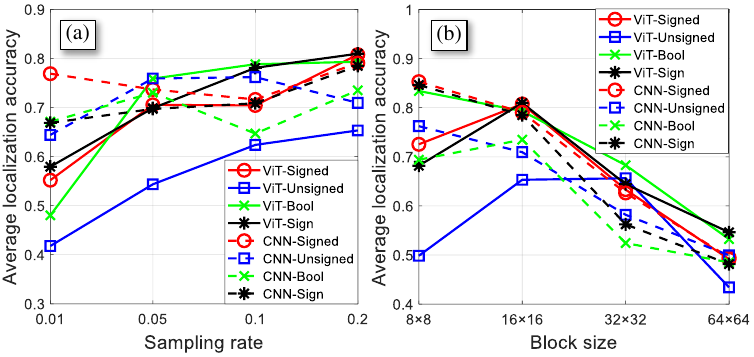}
    \caption{Average localization accuracy vs. (a) sampling rate with block size as 16 and (b) block size with sampling rate as 0.2 under various cases.}
    \label{fig13}
\end{figure}

Fig.~\ref{fig13}(a) illustrates the average localization accuracy versus sampling rate. The results reveal a general trend: localization accuracy improves with increasing sampling rates. This improvement is more stable in ViT, whereas CNN exhibits an oscillatory upward pattern. Consistent with the classification accuracy observations, ViT achieves superior localization performance, particularly when employing binary measurement matrices (i.e., Bool/Sign). Notably, at an extremely low sampling rate ($S_r$=0.01), CNN outperforms ViT. However, ViT rapidly surpasses CNN as the sampling rate increases.

Fig.~\ref{fig13}(b) illustrates the relationship between average localization accuracy and block size. The results demonstrate an overall decline in localization accuracy as block size increases. This degradation occurs primarily because the optical neural network layer computes weighted sums over image blocks, causing larger blocks to lose finer intra-block positional information. Additionally, the reduction in classification accuracy at larger block sizes further impairs localization performance. Consistent with the classification trends, ViT achieves superior localization accuracy, particularly when utilizing binary measurement matrices (Bool/Sign), which yield higher precision.

\subsection{Efficiency Analysis}
The response time of photoelectric sensors is negligible relative to that of DMD, hence, the sampling time of the proposed method yields $T_s=b\cdot m \cdot T_{DMD}$. Adopting typical parameters $b=256$, $m=64$ alongside a 200 kHz working frequency, the corresponding temporal sampling period is merely 82 ms. When an area-scan industrial camera substitutes the discrete photoelectric sensor, FOV parallel projection can replace sliding-window triggering, whereby the total sampling duration is dominated by the sensor's inherent response time $T_{\rm Sensor}$, and the resultant measurement time is simplified to $m\cdot T_{\rm Sensor}$. Given $m=64$ and a typical sensor response time of 100 $\mu s$, the measurement latency is confined to the $ms$ scale, comparable to or superior to conventional industrial-camera-based visual inspection approaches.

Table \ref{tab1} summarizes the efficiency comparisons in various stages. In terms of optical sampling data volume, the proposed method reduces raw data size by 90\% leveraging CS. Computational complexity is quantified via floating-point operations (FLOPs), computed using the Python toolkit $thop$. Partial neural-network computation is physically realized via an optical computing architecture, drastically cutting the overall computational burden, i.e., CNN is reduced by approximately 60\%. Restricted to a single optical convolution layer, the total parameter count remains comparable to baseline schemes. Regarding labeling efficiency, the proposed framework eliminates the time-consuming manual annotation of defect contour required in current image-based methods.

\begin{table}
  \centering
  \large
  \caption{Efficiency Test in Various Stages}
  \label{tab1}
  \resizebox{\linewidth}{!}{
    \begin{tabular}{lllllll}
        \hline
        \multicolumn{1}{c}{{Methods}} & \multicolumn{2}{c}{Optical part} & \multicolumn{2}{c}{Electronic part} & \multicolumn{2}{c}{Dataset annotation}\\
    \cline{2-5}       & Data volume & Time & FLOPs & Parameters & \\
        \hline
        ViT: image & 1048K points & $ms$ level & 5.26G & 0.99M & Category and defect\\
        CNN: image & 1048K points & $ms$ level & 4.04G & 135.24M & contour (in days)\\
        \hline
        ViT: $S_r$=0.1 & 105K points & $ms$ level & 4.85G & 0.89M &  No or only\\
        CNN: $S_r$=0.1 & 105K points & $ms$ level & 1.59G & 134.64M & category\\
        \hline
    \end{tabular}
    } 
\end{table}

\section{Conclusions}
To address the challenges of data bandwidth overload and inefficient shape-level annotation in robotic visual inspection systems, this paper proposes an optoelectronic architecture solution. This architecture integrates the physical sampling of compressive sensing with optical convolutional layers, unifying data acquisition and feature extraction while circumventing complex reconstruction processes. By leveraging the cross-modal alignment capabilities of the CLIP model, the method utilizes semantic text to replace labor-intensive defect shape-level annotation, guiding attention maps to achieve shape-level defect localization. In experiments on tiny detection in transparent materials, the framework achieves a 90\% data reduction when using ViT as the backbone and reduces computational cost by approximately 60\% with CNN as the backbone.

In terms of limitations, it still requires re-collection of data and retraining of models for different industrial inspection scenarios. Future work could enhance the model's generalization across defect shapes by incorporating multi-defect geometries through additional simulation data. Furthermore, the optical path setup in this method is more complex than traditional camera operations. Instrument packaging presents a viable solution to streamline the optical path deployment.



\end{document}